\documentclass{article}
\usepackage{ijcai20}

\usepackage{times}
\usepackage{soul}
\usepackage{url}
\usepackage[hidelinks]{hyperref}
\usepackage[utf8]{inputenc}
\usepackage[small]{caption}
\usepackage{graphicx}
\usepackage[table,xcdraw]{xcolor}
\usepackage{array}
\usepackage{multirow}
\usepackage{amsmath}
\usepackage{amsthm}
\usepackage{booktabs}
\usepackage{algorithm}
\usepackage{algorithmic}
\title{ImmuNetNAS: An Immune-network approach for searching Convolutional Neural Network Architectures}

\author{
Chen Kefan$^1$
\and
Wei Pang$^2$
\affiliations
$^1$First Affiliation\\
$^2$Second Affiliation
\emails
garychenai@gmail.com,
w.pang@hw.ac.uk
}

\begin{document}

\maketitle

\begin{abstract}
  In this research, we propose ImmuNetNAS, a novel Neural Architecture Search (NAS) approach inspired by the immune network theory. The core of ImmuNetNAS is built on the original immune network algorithm, which iteratively updates the population through hypermutation and selection, and eliminates the self-generation individuals that do not meet the requirements through comparing antibody affinity and inter-specific similarity. In addition, in order to facilitate the mutation operation, we propose a novel two-component based neural structure coding strategy. Furthermore, an improved mutation strategy based on Standard Genetic Algorithm (SGA) was proposed according to this encoding method. Finally, based on the proposed two-component based coding method, a new antibody affinity calculation method was developed to screen suitable neural architectures. Systematic evaluations demonstrate that our system have achieved good performance on both the MNIST and CIFAR-10 datasets. We open source our code on GitHub\footnote{\url{ https://github.com/cat-loves-donuts/ImmuNetNAS}} in order to share it with other deep learning researchers and practitioners.
\end{abstract}

\section{Introduction \label{sec:introduction}}

Thanks to manual design by human experts, neural networks have improved rapidly in image and speech recognition in recent years. Considering modern neural architectures tend to be very sophisticated and the limitations of human labour, it is often not sustainable to manually build specific neural networks for any new problems. Therefore, this necessitates Neural Architecture Search (NAS)\cite{elsken2019neural}, an emerging research area.

In this research, we propose a novel NAS strategy called ImmuNetNAS which implemented by Immune Network Algorithm \cite{de2002artificial} for searching the architecture of Convolutional Neural Networks (CNNs) \cite{lecun1998gradient}. In addition, based on Xie's coding method \cite{xie2017genetic}, we further developed a new two-component based coding method. Finally, based on our coding method, we designed and improved a mutation method and interspecific similarity calculation method. At present, to the best of our knowledge, very few researchers have used immune-inspired algorithms on NAS \cite{frachon2019immunecs,barbosa2008evolving,pasti2010neural}, and even fewer researchers tried to search deep neural architectures with artificial immune systems. So another motivation of our work is to explore the potential of immune-inspired algorithms as applied to NAS and assess their effectiveness.

The rest of the paper is organised as follows: Section \ref{sec:related-work} presents related work; Section \ref{sec:ImmuNetNAS} introduces our proposed method ImmuNetNAS; Section \ref{sec:experiments all} reports the experimental results; and Section \ref{sec:conclusion} concludes the paper and explores possible future directions.

\section{Related Work \label{sec:related-work}}

\subsection{Immune Network Algorithm}
Immune Networks Algorithm \cite{de2002artificial} was inspired by the immune network theory proposed by Jerne \cite{jerne1974towards}. Based on this theory, de Castro and Von Zuben \cite{de2002ainet} proposed an immune network algorithm named aiNET which could improve the convergence speed of the population. However, the immune network algorithms still have some issues, such as the large number of B cells which greatly reduces the operation efficiency and increases the network complexity\cite{1121}. Therefore, we need to adapt the original immune network algorithm to meet our specific problem requirements.

\subsection{Neural architecture search (NAS)}
 According to the NAS survey by Elsken $et$ $al$ \cite{elsken2019neural}, there are two aspects, search structure and search strategy, are very important in NAS research.

At present, there are three most commonly used search structures. The first one is the chain structure which is used by Xie $et$ $al$. \cite{xie2017genetic}. The second one is multi-branch structure, such as the structure of DenseNet \cite{huang2017densely}. The last one is hierarchical structure. Many NAS approaches tend to use this structure in order to obtain competitive results, and some of them have outperformed human design \cite{elsken2018efficient,zoph2018learning,real2019regularized,liu2017hierarchical,liu2018progressive}.

Evolutionary algorithms (e.g. genetic algorithms) and reinforcement learning are the two most commonly used search strategies. Many researchers are trying to improve evolutionary algorithms to obtain better results on NAS. For example, Real $et$ $al$. \cite{real2019regularized} added tournament selection, and an age property in original evolutionary algorithm. Elsken $et$ $al$. \cite{elsken2018efficient} used the Lamarckian genetic algorithm to allow the children to get the experiences from their parents. Zoph and Le \cite{zoph2016neural} used a recurrent neural networks to represent the architectures of convolutional neural networks and recurrent neural networks, and discovered competitive architectures by using reinforcement learning.

\subsection{Structure Representation Design}
In evolutionary and immune-inspired algorithms, the mutation operation will significantly affect the performance. Thus, most current studies have developed specific coding strategies to encode the neural network structures for the mutation operations to work on. Liu $et$ $al$. \cite{liu2017hierarchical} proposed a hierarchical encoding method which is useful and has a large enough search space, but relatively complex. On the other hand, Xie $et$ $al$. \cite{xie2017genetic} proposed a binary encoding method, which is simple, but reduces the search space.

\section{ImmuNetNAS \label{sec:ImmuNetNAS}}

In this section we first present the Convolutional Neural Network cells design, and then we propose the encoding method, and finally present the details of ImmuNetNAS design and list the pseudo code.

\subsection{Convolutional Neural Network cells design \label{Convolutional Neural Network cells design}}
ImmuNetNAS uses the hierarchical search strategy whose target is multiple cells to find better structures. At the same time, we use a chain structure mentioned in Section 2.2 to connect different cells and no skipping operation occurs outside the cells. All the skipping operations inside the cells are using the principle of ResNet \cite{huang2017densely}. In the definition of this design, there are 8 layer types for a cell to choose, as shown in Table 1.

\begin{table}[]
\small
\centering
\begin{tabular}{|p{2.5cm}<{\centering}|p{2.3cm}<{\centering}|p{0.8cm}<{\centering}|p{1.0cm}<{\centering}|}
\hline
\textbf{Name and size}   & \textbf{Layer type} & \textbf{Kernel} & \textbf{Padding} \\ \hline
Conv2d(1X1)     & Convolution  & 1X1                  & 0                \\ \cline{1-4} 
Conv2d(3X3)     & Convolution  & 3X3                  & 1                \\ \cline{1-4} 
Conv2d(5X5)     & Convolution  & 5X5                  & 2                \\ \cline{1-4} 
Conv2d(7X7)     & Convolution  & 7X7                  & 3                \\ \cline{1-4} 
AvgPool2d (3X3) & Average Pooling     & 3X3                  & 1                \\ \cline{1-4} 
AvgPool2d (5X5) & Average Pooling     & 5X5                  & 2                \\ \cline{1-4} 
MaxPool2d (3X3) & Max Pooling         & 3X3                  & 1                \\ \cline{1-4} 
MaxPool2d (5X5) & Max Pooling         & 5X5                  & 2                \\ \hline
\end{tabular}
\caption[Table 3.1: The layer types for cell]{\textit{These two table shows 8 different types of layers which can be chosen for our NAS problem.}}
\end{table}

Each convolutionl layer is followed by a batch normalisation layer and a ReLU layer. After each pooling layer, there is another batch normalisation layer. In order to facility design, all convolutionl layers and pooling layers must replace the image size with the original image size by padding operation. Thus, the stride of kernels are 1. Considering the complexity of the code and the hardware resources, the algorithm only incrementally raises the dimensions of the image to 64, and keeps the dimension. The pooling layer only halves its dimension and finally outputs it to the next layer. This solution will help to ensure the dimension consistency between two cells and reduce the algorithm runtime.

In order to train and test each cell, we added the same input and output layers to each cell during the hierarchical search. The input layer includes a Convolution layer which has 1X1 kernel size, 1 stride and no padding, a Batch normalisation layer and a ReLU layer. The output layer is a single linear layer which a specific number of neurons used to obtain classification results. Figure 1 shows the general design of the CNN cell. Furthermore, each cell has a specific default pooling layer which contains a 1X1 convolution layer, a Batch normalisation layer and a ReLU layer before the output. And all the layers that need to be output will go through the default pooling layer (DePooling in Figure 1), so the default pooling layer controls the output of the entire cell. The motivation of this is to prevent the generation of empty cells, which we will introduce in Section 3.3.

\begin{figure}[]
\centering    
\includegraphics[width=0.3\textwidth]{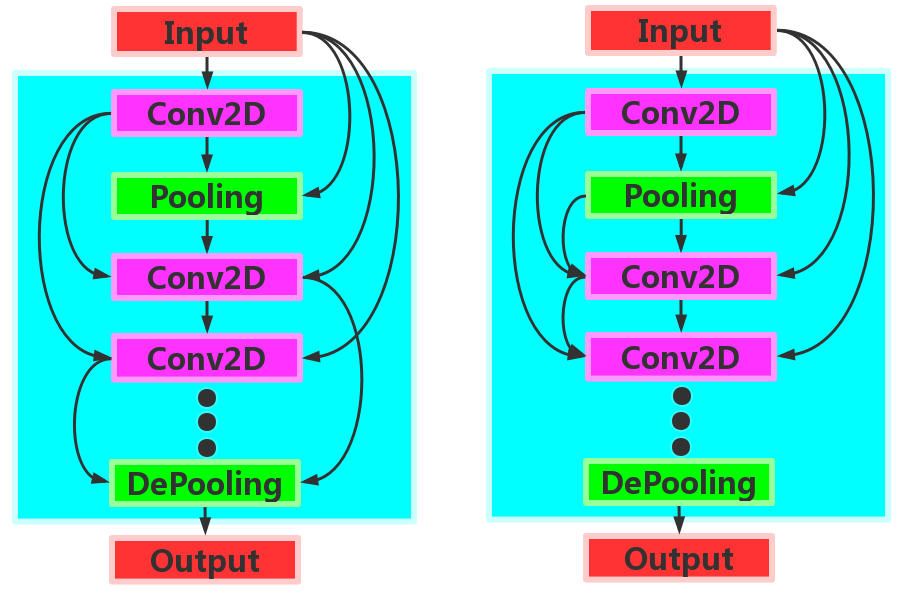}
\caption[Figure 1: The structural design]{\textit{The blue area in the figure represents the cell. The input and output layers outside the cell are created independently. The normal cell structure with multiple skipping connections is shown on the left subfigure. The subfigure on the right is an empty cell structure that no layer is connected with the final DePooling layer. We will introduce a method to prevent empty cells in next section.}}
\label{fig:Figure1}
\end{figure}

\subsection{Encoding Method \label{subsec:Encoding Method}}
Our encoding method is an improvement and further extension of Xie's approach \cite{xie2017genetic}. Our new encoding structure has two components. The first component is used to store the type of each layer of the cell. The second component of the encoding is used to store the connections between the internal layers of the cell. The code and the number of bits $B$ in the second component of a cell $C$ which has $k$ layers are represented in Equations 1 and 2. In Equation 2, $T_{n}$ represents the index of different layer types and $A_{n}$ represents the connection relations between two layers. Those layers that are not connected to the default pooling layer will be discarded during training. Figure 2 shows two examples.

\begin{equation}
   \normalsize B_{C} = \frac{(1 + k)\ast k}{2}
\end{equation}

\begin{equation}
   \normalsize Code =  [ [ T_{1},T_{2},..., T_{k} ], [ A_{1}, A_{2},..., A_{n}, ..., A_{_{B_{C}}} ] ]
\end{equation}

\begin{figure}[] 
\centering    
\includegraphics[width=0.3\textwidth]{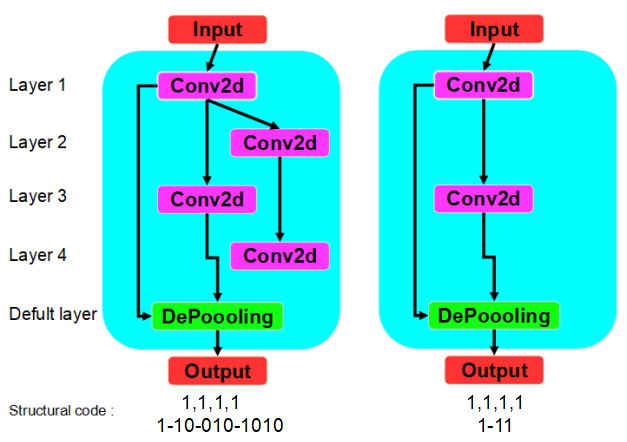}
\caption[Figure 2: The real structure of the cell]{\textit{The left subfigure shows a cell with 4 convolution layers, the right subfigure shows the real structure of the cell. And the second component of structural codes are showing in each sub figure below. Each digit represents whether the previous layers are connected to this layer or not. For the left subfigure, Because Layer 1 is connected to Layer 2, the first digit was 1. And Layer 1 is connected to Layer 3 but Layer 2 is not, so the second and third digits were 10. Only Layer 2 is connected to Layer 3, so the third part of the structural coding was 010. And Layer 1 and Layer 3 connected to DePooling layer, so the last 4 digits were 1010.}}
\label{fig:Figure2}
\end{figure}

With this design, it is possible that empty cells will appear when the system performs the structure search. Therefore, we provide a special mechanism, that is, for the $k$-th layer in the cell, if all the previous layers are not connected to this layer,  the input will be directly connected to this layer by default. This also applies to the last default pooling layer, which means each cell contains at least one default pool layer and no empty cells will be created. The specific method is shown in Fig. 3.

\begin{figure}[] 
\centering    
\includegraphics[width=0.3\textwidth]{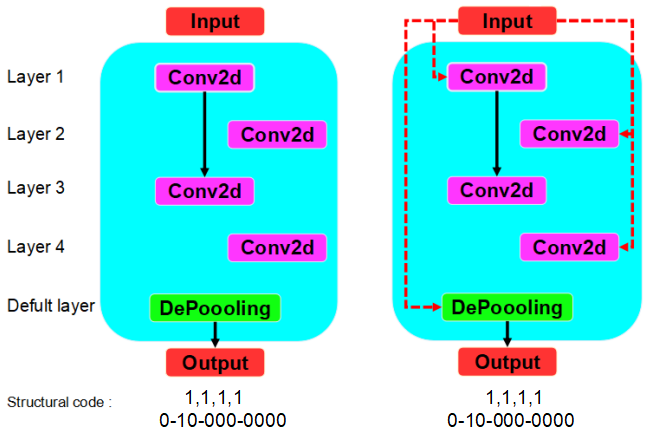}
\caption[Figure 3: The default connection setting]{\textit{The default connection operation. The left subfigure shows a special structure, Layer 1 is connected with Layer 3, and there is not any layer which is connected with the default pooling layer. The right subfigure shows the actual structure of this cell. The red dotted lines represent the default connections, and this cell is equivalent to a neural network with only one 1X1 convolution layer and one pooling layer. And the structural codes are showing below each subfigure.}}
\label{fig:Figure3}
\end{figure}

\subsection{ImmuNetNAS Algorithm Design \label{ImmuNetNAS Algorithm Design}}
Compared with the original immune network algorithms \cite{de2002artificial}, our algorithm is different in the following three aspects:
\begin{itemize}
    \item The target of mutation manipulation is no longer just cells with the highest affinity, but cells selected based on population percentage. 

    \item Network affinity threshold and clone pool threshold were removed. Changing the deletion strategy to remove poorly performed cells based on population size instead of thresholds. On the other hand, a new method for calculating interspecific affinity is proposed.
    
    \item  The system has a independent termination conditions which will help avoid wasting resources and time.
    
\end{itemize}

    The pseudo code of our algorithm is shown in Algorithm 1.

\begin{algorithm}[tb]
\algsetup{linenosize=\tiny}
\scriptsize
\caption{ImmuNetNAS}
\textbf{Input}:$N$: a set of antibodies which represents different models; $h$: the percentage of choosing the highest affinity clones; $a$: the number of new antibodies to be introduced; $q$: the number of cells with the same structure; $G$: the number of generations, $tc$: indicate which stage the system is.
\textbf{Output}:$S$ = a set of memory cells which records the best performed models. 
\begin{algorithmic}[1] 
\REPEAT
\STATE Generate a set of random specific B-cells in $N$.
\STATE $tc \gets 1$
\IF {$tc$ $\geq$ 1}
\FORALL {B-cells $b$ in $N$}
\STATE Connect the previous best performed model in $S$  with $b$.
\ENDFOR
\ENDIF
\STATE $i \gets 0$
\WHILE {$i$ $<$ Generation $G$}
\FORALL {antigens $ag$}
\STATE Calculate the affinity of all B-cells $b$ in $N$ with $ag$.
\ENDFOR
\FORALL{B-cells in $N$}
\STATE Calculate interspecific similarity of each B-cell $b$, place all B-Cells in a temporary set $F$.
\STATE Remove B-cells $b$ in $F$ with high similarity but low antigen affinity, put the rest B-cells back to $N$.
\STATE Select the highest affinity B-cells from $h$ percentage of population, clone them and place clones in $C$.
\ENDFOR
\FORALL {clones $c$ in $C$}
\STATE Mutate $c$ , the degree and number of mutations are inversely proportional to affinity.
\STATE Determine the affinity of $c$ with $ag$.
\STATE Select the highest affinity B-cell clones $c$  from $h$ percentage of population in $C$ and place them in $N$.
\ENDFOR
\FORALL {B-cells in $N$}
\STATE Delete B-cells $b$ in $N$ with low antigen affinity.
\STATE Add randomly generated new B-cells into $N$
\ENDFOR
\STATE $i \gets i + 1$
\ENDWHILE
\STATE Move the best performed B-cell $b$ into the memory cell $S$
\STATE $tc \gets tc + 1$
\UNTIL {The stopping condition has been satisfied}
\end{algorithmic}
\end{algorithm}

\subsubsection{Mutation Method}
We adopt an improved mutation method based on the Standard Genetic Algorithm (SGA) \cite{andre2001improvement}. We added an adaptive mutation rate proposed by Srinvivas $et$ $al$. \cite{srinivas1994adaptive} to the SGA to facilitate ImmuNetNAS jumping out of local optimal solutions. The calculation of mutation rate $P_{m}$ is as follows:

\begin{equation}
\large  P_{m} =\left\{
\begin{array}{lcl}
\frac{k_{1}*(A_{max}-A^{'})}{A_{max}-A_{avg}} & {, A^{'} \geq  A_{avg}}\\
 k_{2} & {, A^{'}<  A_{avg}}\\
\end{array} \right. 
\end{equation}

$k_{1}$ and $k_{2}$ are mutation parameters, $A_{max}$ is the highest affinity in this set of clones, $A_{avg}$ is the average affinity in this set of clones, and $A'$ is the affinity of the target clone. The requirement of this adaptive algorithm is $k_{1} < k_{2}$. The aim is to ensure that better adaptable individuals have more chance to survive, while poorly adapted individuals should undergo more drastic mutations. Based on the coding method, we divides mutations into three different kinds and Figure 4 shows three examples:

\begin{itemize}
\item Light mutations, which will be performed only on the models with the highest affinity. In this case, only the last few bits of the second-dimensional code are mutated.

\item Moderate mutations, which will be performed only on structural models with intermediate affinities. In this case, the entire second dimension code is mutated according to the number of mutation bits previously assigned.

\item Drastic mutations, which will be performed on the model with the lowest affinity. This will mutate the entire code based on the number of mutation bits calculated.
\end{itemize}

\begin{figure}[] 
\centering    
\includegraphics[width=0.4\textwidth]{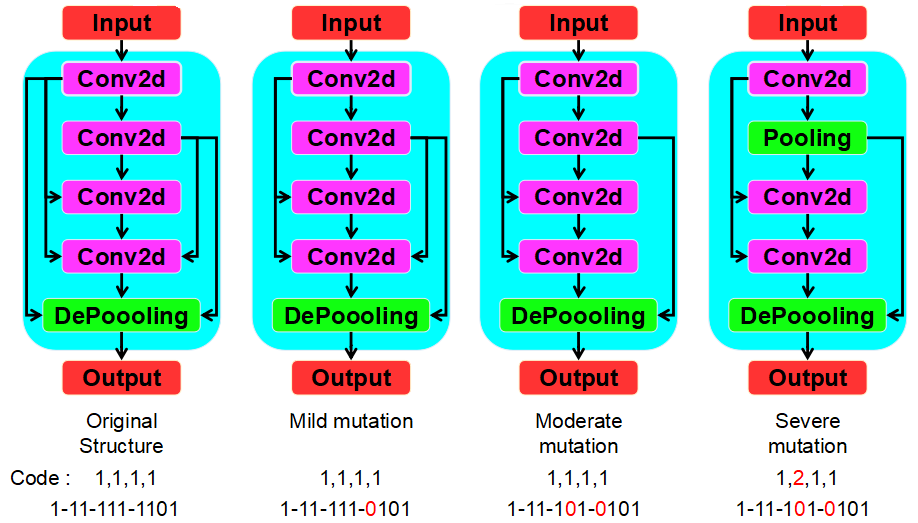}
\caption[Figure 4: Three mutation methods]{\textit{The figure shows the original structure and three different mutation methods. The numbers in red in the code indicate that they have been mutated.}}
\label{fig:Figure4}
\end{figure}

\subsubsection{Interspecific similarity calculation}
One of the commonly used methods for calculating the similarity between different sequences is the Hamming distance. The Hamming distance between two equal-length strings s1 and s2 is defined as the minimum number of substitutions needed to change from one to the other. The Jaccard similarity coefficient $J(A,B)$ and the Tanimoto coefficient $EJ(A,B)$ can be considered as the extensions of the Hamming Distance. These two functions are shown below:  

\begin{equation}
   \normalsize J(A,B) = \frac{\left | A\bigcap B \right |}{\left | A\bigcup  B \right |}
\end{equation}

\begin{equation}
   \normalsize EJ(A,B) = \frac{A*B}{\left \| A \right \|^{2}+\left \| B \right \|^{2} - A*B}
\end{equation}

However, these similarity calculation methods cannot effectively capture the degree of difference between our structural codes. Especially when it comes to binary coding, this situation is particularly prominent. Thus, according to the effect of different bits change on the whole model structure, we firstly compare the similarity of the last $K$ bits of the two structural codes. If the bits of 1 are encountered, it will check the layer type corresponding to this bit, that is, whether the codes in the first dimension are equal. If equal, the corresponding counter increased by 1. Finally, if the quantity difference of the bits of 1 of last $K$ bits between two codes is less than 2, the system will then compare the value of the counter with $S$ which is the total number of 1 in the last $K$ bits. If the value of the counter is greater than or equal to a certain proportion of $S$, the system determines that these two groups of structural codes meet similar conditions, and puts the one with low affinity into the deletion pool.

\subsubsection{Termination conditions}
In ImmuNetNAS, termination conditions are defined as when the accuracy of the best model searched in this generation is not better than previous two generations, the search will be terminated and the best model in the previous two generations will be regarded as the final best model. The reason for this is that the computing resources available to us are not sufficient to support us to do the endless searching.

\section{Experiments \label{sec:experiments all}}
Like other NAS researches, our algorithm requires extensive computing resources. However, due to the fact that limited hardware  resources are available, we only test our algorithm on the MNIST and CIFAR-10 datasets. We used a virtual machine with 8 Intel Xeon CPUs and one NVIDIA Tesla T4 GPU on Google Cloud Platform to perform our experiments.

\subsection{Base frame and Configuration \label{sec:Base frame and Configuration}}
The population is 50, the code ran 20 generations, the Batch size is 16 and using cross entropy as loss function. Because the MNIST dataset is relatively simple compared to the CIFAR-10 dataset, there are some differences on configurations. The total cell number of MNIST dara set is 4, which include 5 layers in each cell. The total cell number of CIFAR-10 dataset is 6, which include 7 layers in each cell. The value of k1 and k2 in the adaptive mutation rate are 0.1 and 0.2 respectively. The percentage in the interspecific similarity calculation was set to 2/3. The details of optimizer are shown in Tables 2. 

\begin{table}[]
\small
\begin{center}
\begin{tabular}{|c|c|c|c|}
\hline
                          & \begin{tabular}[c]{@{}c@{}}Operator\end{tabular} & \multicolumn{2}{c|}{Configuration} \\ \hline
\multirow{2}{*}{MNIST}    & \multirow{2}{*}{SGD}                                                 & learning rate    & momentum        \\ \cline{3-4} 
                      &                                                                      & $10^{-2}$,$10^{-3}$        & 0.9             \\ \hline
\multirow{4}{*}{CIFAR-10} & \multirow{4}{*}{Adam}                                                & learning rate    & beats           \\ \cline{3-4} 
                          &                                                                      & $10^{-3}$             & (0.9,0.999)     \\ \cline{3-4} 
                          &                                                                      & eps              & weight decay    \\ \cline{3-4} 
                          &                                                                      & $10^{-8}$             & $10^{-6}$            \\ \hline
\end{tabular}
\caption[Table 2: Some basic configurations of optimizers]{\textit{The settings of optimizer and loss functions.}}
\end{center}
\end{table}

\subsection{Experiment details \label{sec:experiments}}
\subsubsection{Training methods \label{sec:training methods}}
Currently, the most advanced studies use various training methods. However, due to hardware limitations, we decided to adopt a simpler training method.

For the MNIST, we decided to adopt a training strategy where each cell is trained with only part of the training set so that we can accelerate the training speed. For the CIFAR-10 dataset, we decided to use the complete dataset to train each searched model in each stage once during NAS search. This training strategy can make sure that the model can see the entire data set so that the prediction results are more realistic and valid. For the final training, we will train the whole dataset multiple times to have the model fully trained and get the best accuracy of the model as much as possible.

\subsubsection{Experiments on the MNIST dataset \label{sec:MNIST}}
We conducted two experiments on the MNIST dataset with different learning rate. Our first experiment used a more conservative learning rate of $10^{-3}$ and we ended up with a 92\% accuracy rate. In the second experiment, we adjusted the learning rate to $10^{-2}$, and finally found the model structure with 97\% accuracy. The reason to adjust the learning rate is to give the algorithm a stronger ability to jump out of the local optimal solutions under the current hardware constraints. The two experiments took 42 hours and 46 hours, respectively.

\subsubsection{Experiments on the CIFAR-10 dataset \label{sec:CIFAR-10}}
For CIFAR-10, we conducted three experiments. In the first experiment, we used a total of 4 cells and finally found a model with 65.02\% accuracy. But we suspected that the neural network is too simple, resulting in low accuracy. Thus, we increased the total number of cells to 6, and the accuracy of the model become 81.48\%. In the last experiment, we expanded the search space, increasing the total number of searched models to 3,060 by increasing populations and generations, and finally we found the model with an accuracy of 80.72\%. The time cost and accuracy in the three experiments are shown in Fig. 5.

\begin{figure}[] 
\small
\centering    
\includegraphics[width=0.3\textwidth]{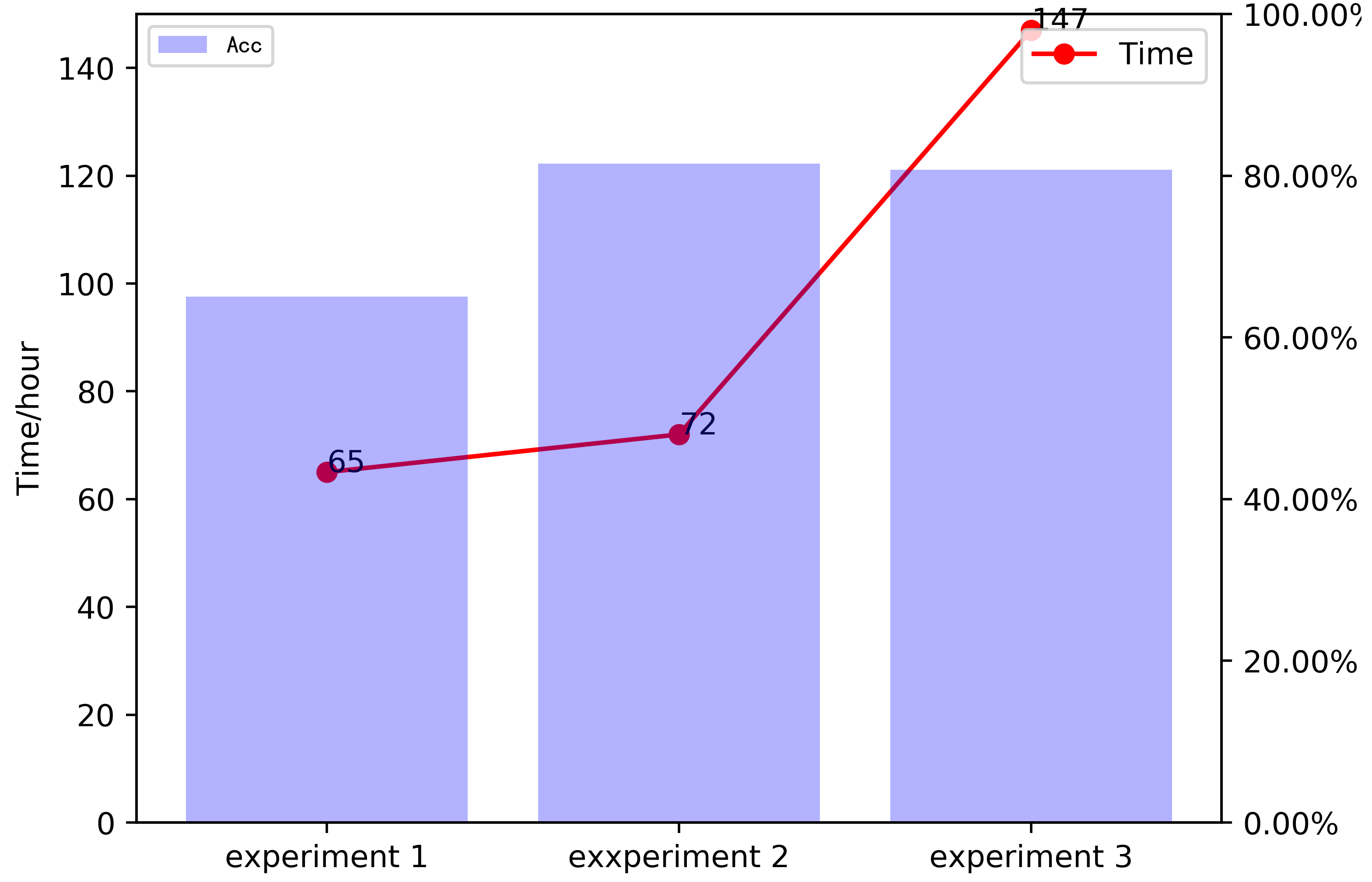}
\caption[Figure 5: CIFAR-10 data set results]{\textit{The results of ImmuNetNAS on the CIFAR-10 dataset.}}
\label{fig:Figure5}
\end{figure}

\section{Results Evaluation \label{sec:evaluation}}

Compared with the relatively simple MNIST dataset, we will focus on analyzing the structure found in the more complex CIFAR-10 dataset.
\subsection{Experiment Analysis on the MINIST Dataset \label{sec:MNIST evaluation}}
The comparisons of our best-performing model and other models on MNIST dataset are showing in Table 3. 

\begin{table}[]
\small
\begin{center}
\begin{tabular}{lr}
\hline
\textbf{Method or Model Name}                                                             & \textbf{Test Error Rate} \\
\hline
Linear Classifiers \footnotemark[2]                                                               & 7.8             \\
Random\footnotemark[2]                                                                           & 1.79            \\
\hline
Neural Nets\footnotemark[2]                                                                   & 0.39            \\

Convolutional Nets\footnotemark[2]                                                               & 0.23            \\
\hline
DeepSwarm Best\\ \cite{byla2019deepswarm}                                                                  & 0.39            \\
DeepSwarm Average\\ \cite{byla2019deepswarm}                                                                & 0.46            \\
\hline
\begin{tabular}[c]{@{}c@{}}ImmuNetNAS\\ (not fully training)
\end{tabular} & 2.7 \\           
\hline
\end{tabular}
\caption[Table 3: The error rate of different models with various methods]{\textit{The error rate of different models on MNIST dataset with various methods.}}
\end{center}
\end{table}
\footnotetext[2]{http://yann.lecun.com/exdb/mnist/}

Compared with other methods, the results of ImmuNetNAS on this experiment are better than those of the linear classifier but far lower than those of other neural network methods. One possible reason might be that the training strategy of our method was that each cell was trained only with a quarter of the entire dataset. Therefore, each cell may not have enough chance to see the full dataset, which may affect the final performance of the model.

\subsection{Experiment Analysis on the CIFAR-10 Dataset \label{sec:MNIST evaluation}}
We analyze the model that we get the best results from 4 aspects: comprehensive situation, hierarchical search, search strategy, and accuracy.

\subsubsection{Comprehensive evaluation}
In our experiments, a satisfactory result was found in the CIFAR-10 dataset, with accuracy reaching 82\%. The actual internal structure of each cell obtained by structural search is shown in Figure 6. Table 4 shows the types of corresponding layers of each cell.

\begin{figure}[]
\centering    
\includegraphics[width=0.4\textwidth]{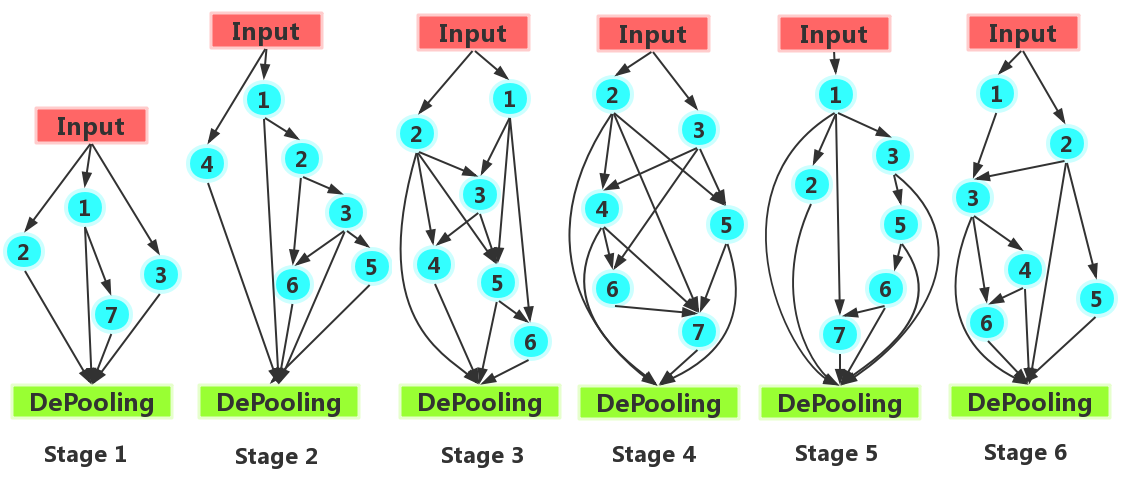}
\caption[Figure 6: The structure of the best performed result]{\textit{The structure of the first 3 stages of Experiment 2.}}
\label{fig:Figure6}
\end{figure}

\begin{table*}[]
\small
\begin{tabular}{|c|c|c|c|c|c|c|c|}
\hline
Cell & Layer 1                             & Layer 2      & Layer 3      & Layer 4                              & Layer 5                              & Layer 6                             & Layer 7                              \\ \hline
1    & AvgPool(5X5)                        & Conv2D(5X5)  & AvgPool(5X5) & \cellcolor[HTML]{FE0000}AvgPool(3X3) & \cellcolor[HTML]{FE0000}AvgPool(3X3) & \cellcolor[HTML]{FE0000}Conv2D(7X7) & AvgPool(3X3)                         \\ \hline
2    & AvgPool(5X5)                        & MaxPool(3X3) & MaxPool(3X3) & Conv2D(7X7)                          & MaxPool(3X3)                         & Conv2D(7X7)                         & \cellcolor[HTML]{FE0000}Conv2D(5X5)  \\ \hline
3    & AvgPool(5X5)                        & Conv2D(7X7)  & MaxPool(3X3) & AvgPool(5X5)                         & Conv2D(5X5)                          & Conv2D(5X5)                         & \cellcolor[HTML]{FE0000}AvgPool(3X3) \\ \hline
4    & \cellcolor[HTML]{FE0000}Conv2D(7X7) & MaxPool(3X3) & MaxPool(3X3) & Conv2D(3X3)                          & MaxPool(3X3)                         & AvgPool(3X3)                        & AvgPool(5X5)                         \\ \hline
5    & Conv2D(5X5)                         & MaxPool(3X3) & Conv2D(3X3)  & \cellcolor[HTML]{FE0000}Conv2D(7X7)  & AvgPool(5X5)                         & AvgPool(5X5)                        & MaxPool(5X5)                         \\ \hline
6    & Conv2D(5X5)                         & Conv2D(1X1)  & AvgPool(3X3) & Conv2D(5X5)                          & Conv2D(1X1)                          & MaxPool(3X3)                        & \cellcolor[HTML]{FE0000}Conv2D(5X5)  \\ \hline 
\end{tabular}
\caption[Table 4: The layer types of different layers]{\textit{The layer types of different layers. The red part indicates that the corresponding layer has been deleted or no longer exists}}
\end{table*}

\subsubsection{Hierarchical search evaluation}\label{Hierarchical_search_evaluation}

The top subfigure in Figure 8 shows the performances of models in each stages. 

\begin{figure}[] 
\centering    
\includegraphics[width=0.5\textwidth]{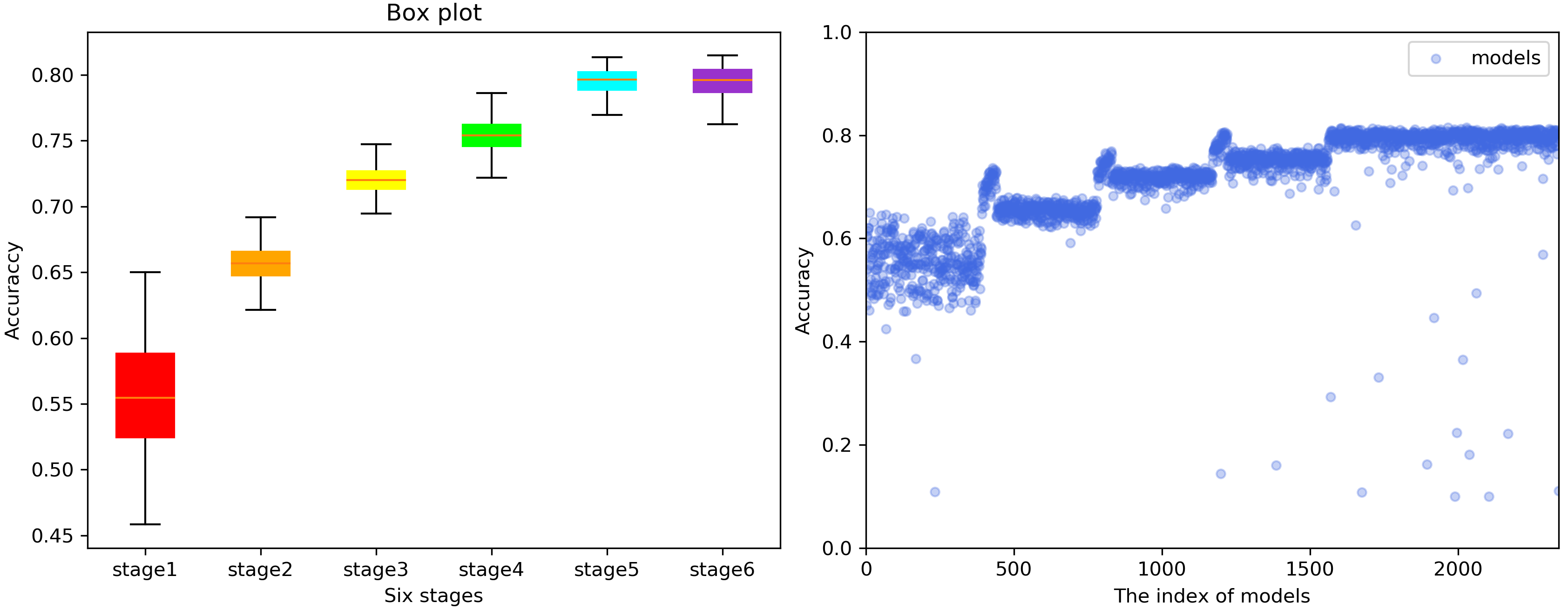}
\caption[Figure 8: The top box graph distribution of all the models searched in experiment 4, the bottom subfigure is scatter plots of 2,340 models]{\textit{The box-plot graph of all the models searched in Experiment 2.}}
\label{fig:Figure8}
\end{figure}

The models produced in the first stage, however, were particularly fragmented. We speculate that this is happening because this is the first cell of NAS searching and it equivalent to a simple and shallow neural network. Their performance on CIFAR-10 dataset can be differ significantly. However, there was also an interesting phenomenon in stage 1: some cells which only have the pooling layers reach more than 60\% test accuracy. We infer that the pooling layer might be stronger than the convolution layer in the extraction of deep features. We will explore this in the future works.

However, in the last two stages, the accuracy of the entire model is not significantly improved even if more cells are spliced. We suspect there are two possible reasons for this. The first one is the design of image size and dimensions. The features of the image might be no longer obvious after processing by multiple cells, so it cannot be extracted and learned. We will focus on improving this design in the future researches. Futhermore, we cannot rule out the possibility that this model is only a locally optimal solution.

\subsubsection{Search strategy evaluation}\label{Search strategy evaluation}
Figure 9 shows the accuracy of all the models searched from the start of NAS to the end. This diagram shows the searched model in each stage, in the order from left to right.

In Stages 3, 4 and 5 of NAS search, a small number of models which were the population randomly initialized at the beginning of each stages were showing better performance than the models which were generated by our mutation method. Although we did not find any errors after reviewing the code, we cannot completely rule out this possibility, so we open sourced our code. There are also some other probabilities which might cause this phenomenon. Firstly, we suspect this situation was caused by the characteristics and mutation degree of the algorithm. Because we designed the mutation function of the algorithm according to the number of the best performed models of a certain population proportion. When the population is initialized, if the randomly generated model has already have the best performance of all the models that can be searched at the current stage, then, no matter how the system mutates new structures, it is impossible to get better performance models. In addition, we suppose that the design of mutation function is not reasonable enough. This makes it difficult to mutate a model that can jump out of the local optimal solutions. 

\subsubsection{Accuracy evaluation\label{Accuracy evaluation}}

Table \ref{Table 5} shows the results of the state-of-art studies.

\begin{table}[]
\small
\centering
\begin{tabular}{lr}
\hline
\textbf{Name}                        & \textbf{Error rate} \\
\hline
Highway Network~\cite{srivastava2015highway}              & 7.72       \\
ResNet-1001 \\+ pre-activation~\cite{he2016identity} & 4.62       \\
DenseNet (k=24)~\cite{huang2017densely}              & 3.74       \\
\hline
NAS v3~\cite{zoph2016neural}                       & 3.65       \\
NASNet-A (7 @ 2304)~\cite{zoph2018learning}          & 2.97       \\
\hline
Genetic CNN~\cite{xie2017genetic}                  & 7.10       \\
Large-scale Evolution~\cite{real2017large}        & 5.4        \\
Evolutionary search~\cite{liu2017hierarchical}          & 3.63       \\
CGP-CNN (ResSet)~\cite{suganuma2017genetic}             & 5.98       \\
\hline
Immune-inspired NAS(not fully train)         & 19         \\  
Immune-inspired NAS(fully train)         & 18           \\
\hline
\end{tabular}
\caption[Table 5: The results of the state-of-art studies]{\textit{The results of the state-of-art studies. }}
\label{Table 5}
\end{table}

Clearly, the models discovered by our system are performing the worst in these studies. However, it cannot be denied that we cannot explore the search space and train the model as much as possible as other studies because of our time limit and insufficient computing resources. But compared to other works we found a relatively good result with less computing resource cost and less time.

However, the results of the full training which we used SGD optimizer and 0.9 momenta did not improve much. Although we tried to change the learning rate, the momentum parameters, and the optimizer, there was no significant improvement in this model. So we would speculate that this structure might be a locally optimal result. This issue would be investigate in the future.

\section{Conclusions \label{sec:conclusion}}

This paper adopts the improved immune network algorithm on neural architecture search tasks. Our basic idea is using a two-component coding structure to encode the structures in each cell, and get neural structures which shows better performance based on our immune-inspired algorithm. In the experiment, we modified the immune network algorithm to make it more suitable for NAS problems, and adopted adaptive mutation parameters to make the algorithm obtain better global search performance. Finally, the neural structures with good performance were found on both MNIST and CIFAR-10 dataset.

Although we made many surprising discoveries during the experiments, our system still had room for improvements. For example, adding a selection probability, for example, and let the algorithm choose to mutate in the direction of better performance like Edvinas $et$ $al$. \cite{byla2019deepswarm}. Furthermore, we will reduce the selection pressure by adding potential parameters to population, which is similar to Real $et$ $al$. \cite{real2018regularized}. On the other hand, we will try to treat NAS tasks as a multi-objective optimize problems(MOP) and use immune-inspired algorithm to simultaneously solve multiple aspects of the NAS problem, such as resource consumption, efficiency, and portability.

\bibliographystyle{named}
\bibliography{ijcai20}

\begin{thebibliography}{}

\bibitem[\protect\citeauthoryear{Andre \bgroup \em et al.\egroup
  }{2001}]{andre2001improvement}
Jerome Andre, Patrick Siarry, and Thomas Dognon.
\newblock An improvement of the standard genetic algorithm fighting premature
  convergence in continuous optimization.
\newblock {\em Advances in engineering software}, 32(1):49--60, 2001.

\bibitem[\protect\citeauthoryear{Barbosa \bgroup \em et al.\egroup
  }{2008}]{barbosa2008evolving}
Bruno~HG Barbosa, Lam~T Bui, Hussein~A Abbass, Luis~A Aguirre, and
  Ant{\^o}nio~P Braga.
\newblock Evolving an ensemble of neural networks using artificial immune
  systems.
\newblock In {\em Asia-Pacific Conference on Simulated Evoluation and
  Learning}, pages 121--130. Springer, 2008.

\bibitem[\protect\citeauthoryear{Byla and Pang}{2019}]{byla2019deepswarm}
Edvinas Byla and Wei Pang.
\newblock Deepswarm: Optimising convolutional neural networks using swarm
  intelligence.
\newblock {\em arXiv preprint arXiv:1905.07350}, 2019.

\bibitem[\protect\citeauthoryear{De~Castro and Timmis}{2002}]{de2002artificial}
L~Nunes De~Castro and Jon Timmis.
\newblock An artificial immune network for multimodal function optimization.
\newblock In {\em Proceedings of the 2002 Congress on Evolutionary Computation.
  CEC'02 (Cat. No. 02TH8600)}, volume~1, pages 699--704. IEEE, 2002.

\bibitem[\protect\citeauthoryear{de Castro and Von~Zuben}{2002}]{de2002ainet}
Leandro~Nunes de~Castro and Fernando~J Von~Zuben.
\newblock ainet: an artificial immune network for data analysis.
\newblock In {\em Data mining: a heuristic approach}, pages 231--260. IGI
  Global, 2002.

\bibitem[\protect\citeauthoryear{Elsken \bgroup \em et al.\egroup
  }{2018}]{elsken2018efficient}
Thomas Elsken, Jan~Hendrik Metzen, and Frank Hutter.
\newblock Efficient multi-objective neural architecture search via lamarckian
  evolution.
\newblock {\em arXiv preprint arXiv:1804.09081}, 2018.

\bibitem[\protect\citeauthoryear{Elsken \bgroup \em et al.\egroup
  }{2019}]{elsken2019neural}
Thomas Elsken, Jan~Hendrik Metzen, and Frank Hutter.
\newblock Neural architecture search: A survey.
\newblock {\em Journal of Machine Learning Research}, 20(55):1--21, 2019.

\bibitem[\protect\citeauthoryear{Frachon \bgroup \em et al.\egroup
  }{2019}]{frachon2019immunecs}
Luc Frachon, Wei Pang, and George~M Coghill.
\newblock Immunecs: neural committee search by an artificial immune system.
\newblock 2019.

\bibitem[\protect\citeauthoryear{He \bgroup \em et al.\egroup
  }{2016}]{he2016identity}
Kaiming He, Xiangyu Zhang, Shaoqing Ren, and Jian Sun.
\newblock Identity mappings in deep residual networks.
\newblock In {\em European conference on computer vision}, pages 630--645.
  Springer, 2016.

\bibitem[\protect\citeauthoryear{Huang \bgroup \em et al.\egroup
  }{2017}]{huang2017densely}
Gao Huang, Zhuang Liu, Laurens Van Der~Maaten, and Kilian~Q Weinberger.
\newblock Densely connected convolutional networks.
\newblock In {\em Proceedings of the IEEE conference on computer vision and
  pattern recognition}, pages 4700--4708, 2017.

\bibitem[\protect\citeauthoryear{Jerne}{1974}]{jerne1974towards}
Niels~K Jerne.
\newblock Towards a network theory of the immune system.
\newblock {\em Ann. Immunol.}, 125:373--389, 1974.

\bibitem[\protect\citeauthoryear{LeCun \bgroup \em et al.\egroup
  }{1998}]{lecun1998gradient}
Yann LeCun, L{\'e}on Bottou, Yoshua Bengio, Patrick Haffner, et~al.
\newblock Gradient-based learning applied to document recognition.
\newblock {\em Proceedings of the IEEE}, 86(11):2278--2324, 1998.

\bibitem[\protect\citeauthoryear{Liu \bgroup \em et al.\egroup
  }{2017}]{liu2017hierarchical}
Hanxiao Liu, Karen Simonyan, Oriol Vinyals, Chrisantha Fernando, and Koray
  Kavukcuoglu.
\newblock Hierarchical representations for efficient architecture search.
\newblock {\em arXiv preprint arXiv:1711.00436}, 2017.

\bibitem[\protect\citeauthoryear{Liu \bgroup \em et al.\egroup
  }{2018}]{liu2018progressive}
Chenxi Liu, Barret Zoph, Maxim Neumann, Jonathon Shlens, Wei Hua, Li-Jia Li,
  Li~Fei-Fei, Alan Yuille, Jonathan Huang, and Kevin Murphy.
\newblock Progressive neural architecture search.
\newblock In {\em Proceedings of the European Conference on Computer Vision
  (ECCV)}, pages 19--34, 2018.

\bibitem[\protect\citeauthoryear{Pasti \bgroup \em et al.\egroup
  }{2010}]{pasti2010neural}
Rodrigo Pasti, Leandro~Nunes de~Castro, Guilherme~Palermo Coelho, and
  Fernando~Jos{\'e} Von~Zuben.
\newblock Neural network ensembles: immune-inspired approaches to the diversity
  of components.
\newblock volume~9, pages 625--653. Springer, 2010.

\bibitem[\protect\citeauthoryear{Real \bgroup \em et al.\egroup
  }{2017}]{real2017large}
Esteban Real, Sherry Moore, Andrew Selle, Saurabh Saxena, Yutaka~Leon Suematsu,
  Jie Tan, Quoc~V Le, and Alexey Kurakin.
\newblock Large-scale evolution of image classifiers.
\newblock In {\em Proceedings of the 34th International Conference on Machine
  Learning-Volume 70}, pages 2902--2911. JMLR. org, 2017.

\bibitem[\protect\citeauthoryear{Real \bgroup \em et al.\egroup
  }{2018}]{real2018regularized}
Esteban Real, Alok Aggarwal, Yanping Huang, and Quoc~V Le.
\newblock Regularized evolution for image classifier architecture search.
\newblock {\em arXiv preprint arXiv:1802.01548}, 2018.

\bibitem[\protect\citeauthoryear{Real \bgroup \em et al.\egroup
  }{2019}]{real2019regularized}
Esteban Real, Alok Aggarwal, Yanping Huang, and Quoc~V Le.
\newblock Regularized evolution for image classifier architecture search.
\newblock In {\em Proceedings of the aaai conference on artificial
  intelligence}, volume~33, pages 4780--4789, 2019.

\bibitem[\protect\citeauthoryear{Srinivas and
  Patnaik}{1994}]{srinivas1994adaptive}
Mandavilli Srinivas and Lalit~M Patnaik.
\newblock Adaptive probabilities of crossover and mutation in genetic
  algorithms.
\newblock {\em IEEE Transactions on Systems, Man, and Cybernetics},
  24(4):656--667, 1994.

\bibitem[\protect\citeauthoryear{Srivastava \bgroup \em et al.\egroup
  }{2015}]{srivastava2015highway}
Rupesh~Kumar Srivastava, Klaus Greff, and J{\"u}rgen Schmidhuber.
\newblock Highway networks.
\newblock {\em arXiv preprint arXiv:1505.00387}, 2015.

\bibitem[\protect\citeauthoryear{Suganuma \bgroup \em et al.\egroup
  }{2017}]{suganuma2017genetic}
Masanori Suganuma, Shinichi Shirakawa, and Tomoharu Nagao.
\newblock A genetic programming approach to designing convolutional neural
  network architectures.
\newblock In {\em Proceedings of the Genetic and Evolutionary Computation
  Conference}, pages 497--504. ACM, 2017.

\bibitem[\protect\citeauthoryear{Timmis and Neal}{2000}]{1121}
J.~Timmis and M.~J. Neal.
\newblock {A Resource Limited Artificial Immune System for Data Analysis}.
\newblock {\em Research and Development in Intelligent Systems XVII}, pages
  182--196, December 2000.
\newblock Proceedings of ES2000, Cambridge, UK.

\bibitem[\protect\citeauthoryear{Xie and Yuille}{2017}]{xie2017genetic}
Lingxi Xie and Alan Yuille.
\newblock Genetic cnn.
\newblock In {\em Proceedings of the IEEE International Conference on Computer
  Vision}, pages 1379--1388, 2017.

\bibitem[\protect\citeauthoryear{Zoph and Le}{2016}]{zoph2016neural}
Barret Zoph and Quoc~V Le.
\newblock Neural architecture search with reinforcement learning.
\newblock {\em arXiv preprint arXiv:1611.01578}, 2016.

\bibitem[\protect\citeauthoryear{Zoph \bgroup \em et al.\egroup
  }{2018}]{zoph2018learning}
Barret Zoph, Vijay Vasudevan, Jonathon Shlens, and Quoc~V Le.
\newblock Learning transferable architectures for scalable image recognition.
\newblock In {\em Proceedings of the IEEE conference on computer vision and
  pattern recognition}, pages 8697--8710, 2018.

\end{thebibliography}

\end{document}